\documentclass{article}
\usepackage[preprint]{neurips_2026}
\usepackage[utf8]{inputenc}
\usepackage[T1]{fontenc}
\usepackage{hyperref}
\hypersetup{hidelinks,hypertexnames=false}
\usepackage{url}
\usepackage{graphicx}
\usepackage{booktabs}
\usepackage{amsfonts}
\usepackage{amsmath}
\usepackage{nicefrac}
\usepackage{microtype}
\usepackage{xcolor}
\usepackage{float}
\usepackage{needspace}
\usepackage{tikz}
\usetikzlibrary{arrows.meta,positioning}
\defcitealias{xiong2020prenorm}{R.~Xiong et~al., 2020}
\defcitealias{xiong2024dcnv4}{Y.~Xiong et~al., 2024}

\title{Where Global Context and Weight Sharing Work in Compact Vision Backbones}
\author{%
  Michael Schroeder \\
  GE Appliances, a Haier company \\
  \texttt{michael.schroeder2@geappliances.com}
}

\begin{document}
\maketitle

\begin{abstract}
Compact convolutional backbones such as EfficientNet achieve strong parameter efficiency through local depthwise separable convolutions, yet two structural questions remain central to hybrid design: \emph{where does global self-attention improve recognition over pure convolution}, and \emph{can repeated blocks share weights without losing representational capacity?} We investigate both questions on ImageNet-1k (50,000-image validation) using controlled EfficientNet-B0 experiments. First, we show that wrapping a classification backbone in multi-scale feature pyramids (BiFPN) fails to outperform a plain MBConv trunk ($69.99\%$--$71.81\%$ vs.\ $72.86\%$ plain B0) until paired with stage attention ($73.15\%$). In contrast, keeping the MBConv trunk intact and inserting lightweight \textbf{zero-gated residual self-attention adapters} at stages $P_3$ ($28\times 28$ windowed), $P_4$ ($14\times 14$ global), and $P_5$ ($7\times 7$ global) reaches \textbf{74.02\%} standalone and outperforms a bit-for-bit matched plain-B0 control by \textbf{+0.954 percentage points} ($73.88\%$ vs.\ $72.92\%$) across an 88-epoch stateless replay schedule, with learned attention amplitude rising monotonically with semantic depth ($27.2\%$ at $P_3 \rightarrow 45.4\%$ at $P_4 \rightarrow 66.3\%$ at $P_5$). Second, we evaluate recurrent weight sharing across repeated MBConv blocks with pass-specific BatchNorm statistics ($+1,536$ parameters per pass). While post-hoc kernel averaging degrades accuracy to $17.03\%$ and single-template stage tying ($-27.6\%$ parameters) plateaus at $66.10\%$, \textbf{alternating two shared block templates ($[B_1, A, B, A]$) in Stage~6} cuts total model parameters by $10.9\%$ ($582\text{K}$ weights) and recovers by $+10.82$ points in a single epoch to \textbf{70.92\%} ($-1.84$ points vs.\ unshared control).
\end{abstract}

\section{Introduction: Targeted Global Context and Recurrent Depth}
\label{sec:intro}

Convolutional neural networks (CNNs) and Vision Transformers~\citep{dosovitskiy2021vit,touvron2021deit} contribute complementary inductive biases to visual recognition. Local $3\times 3$ and $5\times 5$ depthwise convolutions encode translation equivariance and capture high-frequency spatial edges efficiently at high resolution, whereas multi-head self-attention (MHSA)~\citep{srinivas2021botnet,dai2021coatnet} models long-range dependencies across distant object parts once spatial resolution has been reduced. Recent mobile-oriented hybrid architectures such as MobileViT~\citep{mehta2022mobilevit} and EdgeViTs~\citep{pan2022edgevits} have demonstrated that interleaving local convolutions with global or sparse self-attention improves accuracy-efficiency trade-offs on edge devices. Concurrently, research on parameter sharing in one-shot Neural Architecture Search (NAS)~\citep{pham2018enas,liu2019darts} and implicitly recurrent CNNs~\citep{savarese2019recurrent} has established that convolutional layers can reuse weights across subgraphs or sequential passes, raising the question of how best to share parameters across repeated inverted-residual (MBConv) blocks in compact vision backbones.

When modifying a compact backbone such as EfficientNet-B0~\citep{tan2019efficientnet} ($5.32\text{M}$ parameters), untargeted structural modifications can degrade classification accuracy along two axes. First, object detectors routinely employ Bidirectional Feature Pyramid Networks (BiFPN)~\citep{tan2020efficientdet} to fuse multi-scale features for bounding-box localization; however, in whole-image classification, routing features through bidirectional pyramids introduces additional rescaling and projection layers that compete with the primary classification head. Second, while Stages 4, 5, and 6 of EfficientNet-B0 repeat multiple identical-shape MBConv blocks regularized with stochastic depth~\citep{huang2016stochastic}, forcing every repeat within a stage to share a single weight tensor ($[B_1, A, A, A]$) constrains the stage to iterating a single transformation, limiting hierarchical feature refinement~\citep{savarese2019recurrent} in much the same way that aggressive weight sharing induces gradient coupling in one-shot NAS supernets~\citep{pham2018enas,liu2019darts}.

In this study, we evaluate both design questions on ImageNet-1k~\citep{deng2009imagenet} across all 50,000 validation images. We show that \textbf{(1) sparse zero-gated stage attention ($P_3/P_4/P_5$)} yields a verified $+0.954$-point accuracy improvement under bit-for-bit matched training without heavy pyramid necks, and \textbf{(2) alternating two-template weight sharing ($[B_1, A, B, A]$)} in Stage~6 eliminates $10.9\%$ of total model parameters while retaining $4.82$ points higher accuracy than single-template stage tying.

% Visual Infographic 3: Sparse Stage Attention (P3/P4/P5) and Alternating Block Weight Sharing (ABAB)
\begin{figure}[!htbp]
\centering
\resizebox{\linewidth}{!}{%
\begin{tikzpicture}[
  >=Stealth,
  font=\sffamily\footnotesize,
  stagebox/.style={draw=black!65, rounded corners=3pt, inner sep=0pt, minimum width=1.26cm, minimum height=1.26cm, fill=blue!7},
  attnbox/.style={draw=orange!75!black, very thick, rounded corners=3pt, inner sep=0pt, minimum width=1.26cm, minimum height=1.26cm, fill=orange!14},
  wscard/.style={draw=black!65, rounded corners=4pt, inner sep=0pt, minimum width=3.68cm, minimum height=3.30cm}
]

% ==================== TOP PANEL: SPARSE ZERO-GATED ATTENTION ====================
\node[anchor=west, font=\sffamily\small\bfseries, text=black!85] at (0, 5.55)
  {A. Sparse Zero-Gated Stage Attention: Deep Stages Recruit Global Context (+0.954 pp)};

\node[stagebox, anchor=north west] (in) at (0, 5.15) {%
  \parbox[c][1.10cm][c]{1.16cm}{\centering\footnotesize\textbf{Input}\\$224^2$\\RGB}%
};
\node[stagebox, anchor=north west] (s12) at (1.505, 5.15) {%
  \parbox[c][1.10cm][c]{1.16cm}{\centering\footnotesize\textbf{Stem \&}\\Stages\\1--2}%
};
\node[stagebox, anchor=north west] (s3) at (3.01, 5.15) {%
  \parbox[c][1.10cm][c]{1.16cm}{\centering\footnotesize\textbf{Stage 3}\\MBConv\\$28\times 28$}%
};
\node[attnbox, anchor=north west] (p3) at (4.515, 5.15) {%
  \parbox[c][1.10cm][c]{1.16cm}{\centering\footnotesize\textbf{\color{orange!65!black}P3 Gate}\\Win-Attn\\\textbf{27\%}}%
};
\node[stagebox, anchor=north west] (s45) at (6.02, 5.15) {%
  \parbox[c][1.10cm][c]{1.16cm}{\centering\footnotesize\textbf{Stg 4--5}\\MBConv\\$14\times 14$}%
};
\node[attnbox, anchor=north west] (p4) at (7.525, 5.15) {%
  \parbox[c][1.10cm][c]{1.16cm}{\centering\footnotesize\textbf{\color{orange!65!black}P4 Gate}\\Global\\\textbf{45\%}}%
};
\node[stagebox, anchor=north west] (s67) at (9.03, 5.15) {%
  \parbox[c][1.10cm][c]{1.16cm}{\centering\footnotesize\textbf{Stg 6--7}\\MBConv\\$7\times 7$}%
};
\node[attnbox, anchor=north west] (p5) at (10.535, 5.15) {%
  \parbox[c][1.10cm][c]{1.16cm}{\centering\footnotesize\textbf{\color{orange!65!black}P5 Gate}\\Global\\\textbf{66\%}}%
};

\draw[->, thick, draw=black!65] (in.east) -- (s12.west);
\draw[->, thick, draw=black!65] (s12.east) -- (s3.west);
\draw[->, thick, draw=orange!75!black] (s3.east) -- (p3.west);
\draw[->, thick, draw=orange!75!black] (p3.east) -- (s45.west);
\draw[->, thick, draw=orange!75!black] (s45.east) -- (p4.west);
\draw[->, thick, draw=orange!75!black] (p4.east) -- (s67.west);
\draw[->, thick, draw=orange!75!black] (s67.east) -- (p5.west);

\node[draw=orange!65!black, fill=orange!6, rounded corners=3pt, inner sep=0pt, minimum width=11.80cm, minimum height=1.18cm, anchor=north west] at (0, 3.68) {%
  \parbox[c][0.96cm][c]{11.40cm}{\small\raggedright
    \textbf{\color{orange!65!black}Key Finding:} Zero-initialized gates ($\alpha_s=0$) preserve early convolutional learning while deep stages recruit global context (\textbf{27.2\%} $\rightarrow$ \textbf{45.4\%} $\rightarrow$ \textbf{66.3\%} RMS), yielding \textbf{+0.954 pp} matched gain (\textbf{73.88\%} vs.\ \textbf{72.92\%}; \textbf{74.02\%} standalone).
  }%
};

% ==================== BOTTOM PANEL: ALTERNATING WEIGHT SHARING (ABAB) ====================
\node[anchor=west, font=\sffamily\small\bfseries, text=black!85] at (0, 2.12)
  {B. Stage 6 Recurrent Weight Sharing: Alternating Pairs ($[B_1, A, B, A]$) vs.\ Single Template};

% Left Box: Unshared Baseline
\node[wscard, fill=gray!6, draw=gray!70!black, anchor=north west] (w1) at (0, 1.72) {%
  \parbox[c][3.00cm][c]{3.42cm}{\footnotesize\raggedright
    \textbf{\small\color{black!85}1. Unshared Baseline}\\[4pt]
    \texttt{[B1]$\rightarrow$[B2]$\rightarrow$[B3]$\rightarrow$[B4]}\\[4pt]
    $\bullet$ \textbf{Params:} 5.32M (100\%)\\[3pt]
    $\bullet$ \textbf{Zero-Shot:} 72.87\%\\[3pt]
    $\bullet$ \textbf{1-Ep:} \textbf{72.76\%} (control)\\[4pt]
    \textcolor{black!75}{Independent block weights.}
  }%
};

% Middle Box: Single-Template Tying (Collapsed)
\node[wscard, fill=red!6, draw=red!65!black, anchor=north west] (w2) at (4.06, 1.72) {%
  \parbox[c][3.00cm][c]{3.42cm}{\footnotesize\raggedright
    \textbf{\small\color{red!70!black}2. Single Template}\\[4pt]
    \texttt{[B1]$\rightarrow$[\color{red!70!black}A\color{black}]$\rightarrow$[\color{red!70!black}A\color{black}]$\rightarrow$[\color{red!70!black}A\color{black}]}\\[4pt]
    $\bullet$ \textbf{Params:} 3.85M (-27.6\%)\\[3pt]
    $\bullet$ \textbf{Zero-Shot:} 47.86\%\\[3pt]
    $\bullet$ \textbf{1-Ep:} \textbf{66.10\%} ($-6.66$ pp)\\[4pt]
    \textcolor{red!65!black}{Single template limits depth.}
  }%
};

% Right Box: Alternating ABAB Pair (Sweet Spot)
\node[wscard, fill=teal!7, draw=teal!75!black, very thick, anchor=north west] (w3) at (8.12, 1.72) {%
  \parbox[c][3.00cm][c]{3.42cm}{\footnotesize\raggedright
    \textbf{\small\color{teal!60!black}3. Alternating Pair $\star$}\\[4pt]
    \texttt{[B1]$\rightarrow$[\color{teal!70!black}A\color{black}]$\rightarrow$[\color{purple!75!black}B\color{black}]$\rightarrow$[\color{teal!70!black}A\color{black}]}\\[4pt]
    $\bullet$ \textbf{Params:} 4.74M (-10.9\%)\\[3pt]
    $\bullet$ \textbf{Zero-Shot:} 60.10\%\\[3pt]
    $\bullet$ \textbf{1-Ep:} \textbf{70.92\%} (+10.8 pp)\\[4pt]
    \textcolor{teal!55!black}{Two templates + per-pass BN.}
  }%
};

\end{tikzpicture}%
}
\caption{\textbf{Architectural efficiency through targeted global context and alternating weight sharing.} \textbf{(A)} Each residual adapter $y_s = x_s + \alpha_s \cdot \mathrm{Proj}_s(\mathrm{MHSA}_s(\mathrm{LN}(x_s)))$ initializes at $\alpha_s=0$ so early convolutional feature learning is undisturbed. By epoch 80, learned branch update-to-input RMS rises monotonically with semantic depth ($27.2\%$ at $P_3$ windowed MHSA $\rightarrow 45.4\%$ at $P_4$ global MHSA $\rightarrow 66.3\%$ at $P_5$ global MHSA). Under bit-for-bit identical stateless data replay across 88 epochs, Sparse Attention outperforms Plain-B0 by $+0.954$ percentage points ($73.88\%$ vs.\ $72.92\%$; standalone recovery reaches $74.02\%$). \textbf{(B)} In Stage~6 (the largest parameter stage in EfficientNet-B0), post-hoc kernel averaging degrades zero-shot accuracy to $17.03\%$ and tying all repeated blocks in Stages 4--6 to a single template ($[B_1, A, A, A]$) plateaus at $66.10\%$ ($-6.66$ pp vs.\ control), whereas alternating two shared templates ($[B_1, A, B, A]$) with pass-specific BatchNorm statistics reduces total model parameters by $10.9\%$ ($582\text{K}$ weights) and recovers to $70.92\%$ in a single continuation epoch.}
\label{fig:attention-and-weight-sharing}
\end{figure}
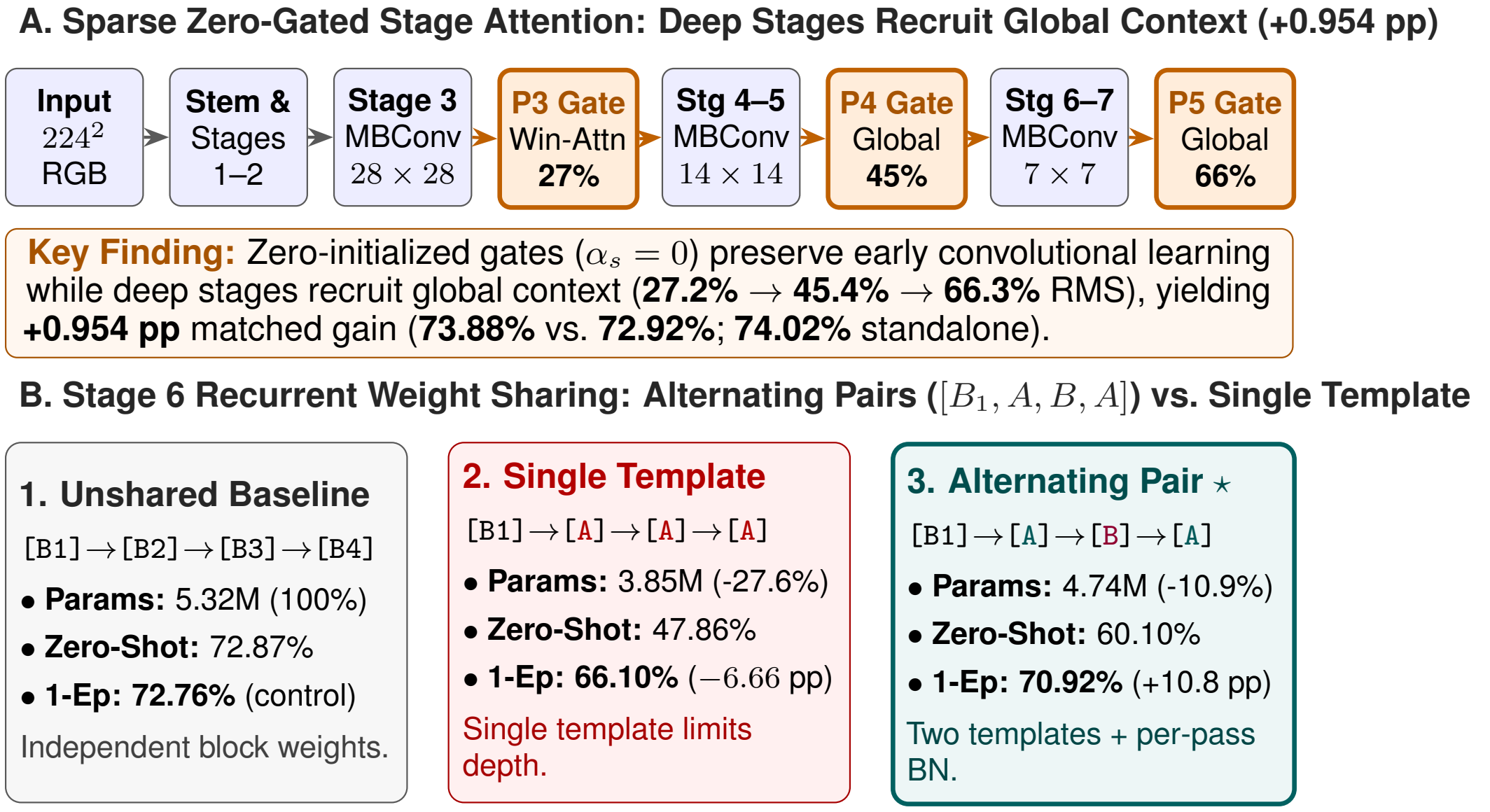

\section{Baseline Foundation and Multi-Scale BiFPN Routing}
\label{sec:baselines-bifpn}

\begin{figure}[!htbp]
\centering
\includegraphics[width=0.95\linewidth]{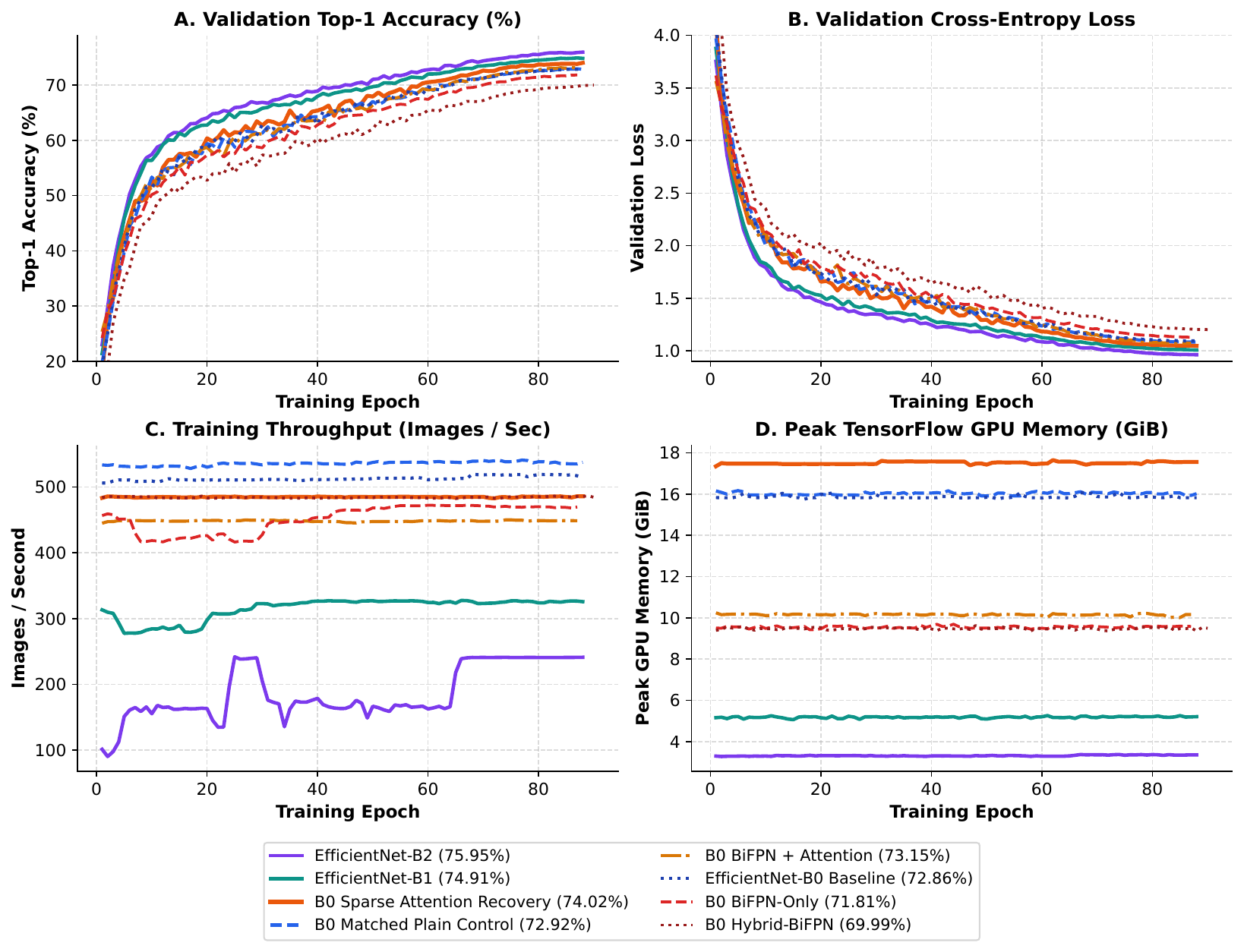}
\caption{\textbf{Full-validation accuracy and resource trajectories across B0--B2 baselines, BiFPN hybrids, and Sparse Stage Attention.} While BiFPN-only variants plateau below the plain B0 baseline ($69.99\%$--$71.81\%$), Sparse Stage Attention ($P_3/P_4/P_5$) achieves the strongest B0-scale trajectory ($74.02\%$ standalone; $73.88\%$ vs.\ $72.92\%$ in the matched pair).}
\label{fig:validation-progress}
\end{figure}

To ground our architectural comparisons, we trained reference EfficientNet-style B0 ($224\times 224$), B1 ($240\times 240$), and B2 ($260\times 260$) backbones for 88 epochs using a standardized ReLU6/SGD recipe, alongside three multi-scale BiFPN~\citep{tan2020efficientdet} classification hybrids (Table~\ref{tab:architecture-summary} and Figure~\ref{fig:validation-progress}):

\begin{table}[!htbp]
\centering
\caption{\textbf{Full 50,000-image ImageNet-1k accuracy across reference scales, BiFPN hybrids, and Sparse Stage Attention.} Under bit-for-bit matched 88-epoch training, Sparse Stage Attention ($P_3/P_4/P_5$) beats Plain-B0 by $+0.954$ percentage points ($73.88\%$ vs.\ $72.92\%$) and reaches $74.02\%$ in standalone recovery.}
\label{tab:architecture-summary}
\footnotesize
\setlength{\tabcolsep}{3.6pt}
\resizebox{\linewidth}{!}{%
\begin{tabular}{@{}llrrrl@{}}
\toprule
\textbf{Cohort / Group} & \textbf{Model Architecture} & \textbf{Input} & \textbf{Sel.\ Ep} & \textbf{Top-1 / Top-5 (\%)} & \textbf{Key Architectural Comparison} \\
\midrule
\textbf{Scale Baselines} & Plain EfficientNet-B0 (A4 Ref) & $224^2$ & e86 / 88 & 72.86 / 91.11 & Primary B0 reference ($5.32\text{M}$ params) \\
\textbf{(88-Epoch Ref)} & Plain EfficientNet-B1 (A7 Ref) & $240^2$ & e87 / 88 & 74.91 / 92.19 & Compound scale ($7.85\text{M}$ params) \\
 & Plain EfficientNet-B2 (A5 Ref) & $260^2$ & e88 / 88 & 75.95 / 92.75 & Largest reference scale ($9.17\text{M}$ params) \\
\midrule
\textbf{BiFPN Hybrids} & B0 + Fresh BiFPN Neck (A2) & $224^2$ & e90 / 90 & 69.99 / 89.29 & Warm trunk + fresh BiFPN lags plain B0 \\
 & B0 + BiFPN Joint Recovery (A3) & $224^2$ & e87 / 87 & 71.81 / 90.38 & Joint BiFPN trails plain B0 ($-1.05$ pp) \\
 & B0 + BiFPN + Stage Attention (A1) & $224^2$ & e86 / 87 & 73.15 / 91.28 & Adding attention lifts BiFPN by $+1.34$ pp \\
\midrule
\textbf{Sparse Attention} & \textbf{Sparse Attention B0 Standalone} & $224^2$ & e88 / 88 & \textbf{74.02 / 91.74} & \textbf{Best B0-scale accuracy ($+1.16$ pp vs.\ A4)} \\
\textbf{Matched Pair} & Plain-B0 Matched Control & $224^2$ & e87 / 88 & 72.92 / 91.15 & Stateless bit-for-bit 88-epoch control \\
\textbf{(Bit-for-Bit)} & \textbf{Sparse Attention B0 Matched} & $224^2$ & e86 / 88 & \textbf{73.88 / 91.68} & \textbf{Verified $+0.954$ pp causal gain ($+0.914$ at e88)} \\
\bottomrule
\end{tabular}%
}
\end{table}

As shown in Table~\ref{tab:architecture-summary}, adding a BiFPN neck alone did not improve ImageNet classification over a plain MBConv trunk ($69.99\%$ when attached fresh to a hybrid trunk; $71.81\%$ when trained jointly for 87 epochs vs.\ $72.86\%$ for plain B0). Only when multi-head self-attention was added alongside BiFPN did accuracy reach $73.15\%$ ($+1.34$ points over no-attention BiFPN). This observation motivated removing the feature pyramid neck altogether to test whether sparse stage attention inserted directly inside the MBConv trunk is the primary contributor to the accuracy improvement.

\section{Sparse Zero-Gated Stage Attention ($P_3/P_4/P_5$)}
\label{sec:sparse-attention}

Prior hybrid architectures integrate self-attention by replacing entire convolutional stages or bottleneck blocks: BoTNet~\citep{srinivas2021botnet} substitutes spatial convolutions in the final $c_5$ stage with global MHSA, CoAtNet~\citep{dai2021coatnet} stacks full Transformer stages across $S_3$ and $S_4$, MobileViT~\citep{mehta2022mobilevit} unfolds spatial feature maps into patches for global Transformer blocks, and EdgeViTs~\citep{pan2022edgevits} alternate local-global-local blocks across stages. Rather than replacing MBConv blocks or re-architecting the trunk, our \textbf{Sparse Attention-Convolution B0} design preserves the exact MBConv and Squeeze-and-Excitation (SE) trunk of EfficientNet-B0 and inserts three lightweight residual attention adapters at the outputs of Stage~3 ($P_3$, $28\times 28$ resolution, $40$ channels), Stage~5 ($P_4$, $14\times 14$ resolution, $112$ channels), and Stage~7 ($P_5$, $7\times 7$ resolution, $320$ channels). To keep quadratic attention cost bounded at $28\times 28$, $P_3$ uses $7\times 7$ windowed self-attention, whereas $P_4$ and $P_5$ use global multi-head self-attention. Each adapter applies pre-layer normalization (Pre-LN)~[\citetalias{xiong2020prenorm}], multi-head self-attention, a linear projection, and a \textbf{zero-initialized learnable gate} $\alpha_s$:
\[
  y_s = x_s + \alpha_s \cdot \mathrm{Proj}_s\!\left(\mathrm{MHSA}_s(\mathrm{LN}(x_s))\right), \qquad \alpha_s(0) = 0.
\]
Initializing $\alpha_s = 0$ ensures that at step~0 the network is functionally identical to plain EfficientNet-B0, preventing uncalibrated attention gradients from perturbing early convolutional feature learning.

\subsection{Bit-for-Bit Matched Stateless Verification ($+0.954$ pp)}
\begin{figure}[!htbp]
\centering
\includegraphics[width=0.95\linewidth]{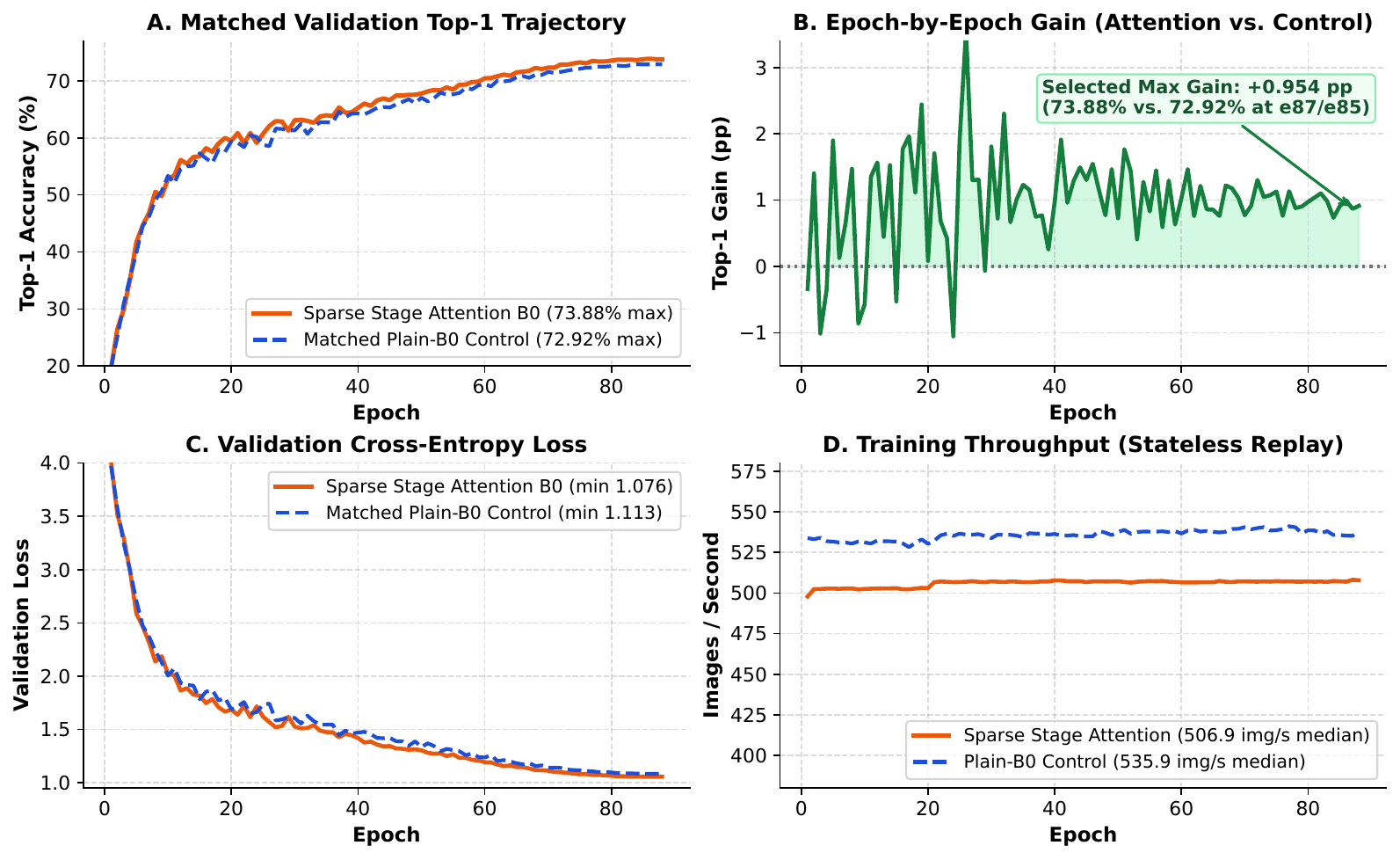}
\caption{\textbf{Bit-for-bit matched 88-epoch comparison of Sparse Stage Attention ($P_3/P_4/P_5$) vs.\ Plain-B0.} Both arms consume identical stateless training streams. Sparse Attention maintains a consistent $\sim 1.0$-point lead throughout convergence, finishing at $73.88\%$ vs.\ $72.92\%$ ($+0.954$ pp).}
\label{fig:matched-attention-pair}
\end{figure}

To isolate the causal contribution of the $P_3/P_4/P_5$ attention adapters without confounding differences in data order or initialization, we executed a strictly paired 88-epoch experiment in which Plain-B0 and Sparse Attention B0 started from the same backbone initialization and consumed \textbf{bit-for-bit identical stateless image batches} across all 112.73 million training presentations (Figure~\ref{fig:matched-attention-pair}). Across the validation trajectory, Sparse Attention maintained a consistent accuracy advantage over Plain-B0 at every full-validation checkpoint from epoch 60 through 88, leading by $+1.216$ percentage points at epoch 63 ($71.102\%$ vs.\ $69.886\%$), by \textbf{$+0.954$ points at the selected checkpoints} ($73.878\%$ at e86 vs.\ $72.924\%$ at e87), and by \textbf{$+0.914$ points at terminal epoch 88} ($73.794\%$ vs.\ $72.880\%$). This accuracy gain incurred only modest computational overhead: across the final 10 epochs (e79--e88) on matched single-GPU hardware, adding the three sparse attention gates reduced training throughput by $5.5\%$ ($479.8$ vs.\ $507.6\text{ images/s}$) and increased peak memory allocation by $9.2\%$ ($17.29\text{ GiB}$ vs.\ $15.83\text{ GiB}$).

\subsection{Learned Gate Activity Scales Monotonically with Semantic Depth}
To understand how the network allocates capacity between local convolutions and self-attention across scales, we measured the ratio of the gated attention update RMS to the trunk input RMS ($\|\alpha_s \Delta x_s\|_{\text{RMS}} / \|x_s\|_{\text{RMS}}$) across a fixed class-balanced 2,000-image validation subset. The learned adapters exhibit a consistent depth-dependent trend: at \textbf{$P_3$ ($28\times 28$ resolution)}, where features primarily encode mid-level edges and local textures, relative branch activity converges to a modest \textbf{27.2\%} (with gating scalar $|\alpha_{P_3}| = 0.258$). At \textbf{$P_4$ ($14\times 14$ resolution)}, where features represent intermediate object part assemblies, relative branch activity rises to \textbf{45.4\%} ($|\alpha_{P_4}| = 0.451$). By \textbf{$P_5$ ($7\times 7$ resolution)}, where spatial dimensions are coarse and channels capture whole-object semantic context, relative branch activity reaches \textbf{66.3\%} ($|\alpha_{P_5}| = 0.720$). Gradient descent thus naturally keeps $P_3$ dominated by local $3\times 3$ and $5\times 5$ depthwise convolutions while recruiting progressively stronger global self-attention in $P_4$ and $P_5$, providing direct stage-level empirical support for the late-attention design philosophy used in modern hybrid vision architectures~\citep{dai2021coatnet,srinivas2021botnet,mehta2022mobilevit,pan2022edgevits}.

\section{Recurrent Weight Sharing: Alternating Pairs ($ABAB$) vs.\ Single Templates}
\label{sec:weight-sharing}

Over $75\%$ of EfficientNet-B0's parameters reside in Stages 4, 5, and 6, each of which begins with a stride-changing entry block ($B_1$) followed by 2 or 3 identical-shape repeat blocks ($B_2, B_3, B_4$). Stage~6 alone contains four $192$-channel MBConv6 blocks ($5\times 5$ depthwise kernels), where each repeated block carries $582,384$ convolutional weights. Parameter sharing across convolutional operations has been widely studied in one-shot Neural Architecture Search, where ENAS~\citep{pham2018enas} and DARTS~\citep{liu2019darts} share kernel weights across candidate edges in a directed acyclic graph, and in implicitly recurrent CNNs, where \citet{savarese2019recurrent} observed that layers learning soft parameter sharing within Wide ResNet stages naturally converge toward 2-layer alternating recurrent loops while keeping transition blocks distinct.

We tested whether repeated MBConv blocks within EfficientNet-B0 can similarly share convolutional and Squeeze-and-Excitation kernels while maintaining \textbf{separate BatchNorm running statistics and affine scales per repeat pass} (since feature distributions evolve across sequential residual steps). Crucially, keeping BatchNorm layers pass-specific introduces negligible parameter and memory overhead: each pass-specific BatchNorm set contributes only $+1,536$ output-channel parameters ($\sim 6\text{ KB}$ in FP32, or $<0.03\%$ of the $5.32\text{M}$-parameter backbone), compared to the $582,384$ convolutional and Squeeze-and-Excitation weights saved per shared Stage~6 block. Starting from our mature B0 checkpoint ($72.868\%$ FP32), we evaluated zero-shot weight sharing (with and without 1,000-image BatchNorm recalibration) followed by a matched \textbf{one-epoch GPU continuation} (5,004 steps at batch size 256; 1.28M examples) across four structural arms (Table~\ref{tab:weight-sharing}):

\begin{table}[!htbp]
\centering
\caption{\textbf{Full 50,000-image ImageNet-1k results for recurrent block weight sharing from mature EfficientNet-B0.} While naive weight averaging degrades accuracy to $17.03\%$ and single-template stage tying ($[B_1, A, A, A]$ across Stages 4--6) plateaus at $66.10\%$, alternating two shared templates in Stage~6 ($[B_1, A, B, A]$) reduces total model parameters by $10.9\%$ and recovers by $+10.82$ points in a single epoch to \textbf{70.92\%}.}
\label{tab:weight-sharing}
\footnotesize
\setlength{\tabcolsep}{3.5pt}
\resizebox{\linewidth}{!}{%
\begin{tabular}{@{}llrrrrrr@{}}
\toprule
\textbf{Weight-Sharing Architecture} & \textbf{Stage 6 Pattern} & \textbf{Params} & \textbf{Param $\Delta$} & \textbf{Zero-Shot} & \textbf{+ 1K BN} & \textbf{1-Ep Cont.} & \textbf{$\Delta$ vs.\ Ctrl} \\
\midrule
Unshared FP32 Baseline / Control & $[B_1, B_2, B_3, B_4]$ & 5.320M & $0.0\%$ & 72.87\% & 70.52\% & \textbf{72.76\%} & $0.00$ pp \\
\midrule
Stages 4--6 Single-Template (Avg Init) & $[B_1, \bar{B}, \bar{B}, \bar{B}]$ & 3.851M & $-27.6\%$ & 0.72\% & 17.03\% & --- & --- \\
Stages 4--6 Single-Template (Copy $B_2$) & $[B_1, A, A, A]$ & 3.851M & $-27.6\%$ & 7.36\% & 47.86\% & 66.10\% & $-6.66$ pp \\
Stage 6 Single-Template (Avg Init) & $[B_1, \bar{B}, \bar{B}, \bar{B}]$ & 4.155M & $-21.9\%$ & 14.84\% & 52.53\% & --- & --- \\
\textbf{Stage 6 Alternating Pair (4-Block)} & \textbf{$[B_1, A, B, A]$} & \textbf{4.738M} & \textbf{$-10.9\%$} & \textbf{28.34\%} & \textbf{60.10\%} & \textbf{70.92\%} & \textbf{$-1.84$ pp} \\
Stage 6 Alternating Pair (5-Block Depth) & $[B_1, A, B, A, B]$ & 4.739M & $-10.9\%$ & 14.37\% & 58.49\% & 70.83\% & $-1.93$ pp \\
\bottomrule
\end{tabular}%
}
\end{table}

The results in Table~\ref{tab:weight-sharing} clarify three structural requirements for weight sharing in compact vision backbones. First, trained convolutional kernels should not be combined by arithmetic averaging ($\bar{B} = \frac{1}{K}\sum_k B_k$): because intermediate channel permutations and signs differ across sequential residual steps, averaging weights across blocks disrupts spatial filter alignment and reduces accuracy to $17.03\%$ even after BatchNorm recalibration. Initializing shared templates by copying the existing block ($A=B_2$) and recalibrating pass-specific BatchNorm statistics on 1,000 images outperforms kernel averaging by \textbf{$+30.83$ percentage points} across Stages 4--6 ($47.86\%$ vs.\ $17.03\%$) prior to any gradient updates.

Second, alternating two shared templates ($[B_1, A, B, A]$) substantially outperforms single-template tying ($[B_1, A, A, A]$). When all repeated blocks in Stages 4--6 are tied to a single template per stage, one epoch of GPU continuation lifts accuracy from $47.86\%$ to $66.10\%$, leaving a $-6.66$-point deficit relative to the unshared control ($72.76\%$). By contrast, alternating two templates in Stage~6 ($[B_1, A, B, A]$, where Block~4 reuses Block~2's weights $A$ while Block~3 retains $B$) eliminates $582,384$ parameters ($10.94\%$ of the model), reaches $60.10\%$ after zero-shot BN recalibration, and recovers to \textbf{70.92\%} after a single continuation epoch—outperforming single-template tying by $+4.82$ points and closing $85.5\%$ of the initial $12.66$-point tying gap within just 5,004 gradient updates. Consistent with the two-layer recurrence observed in Wide ResNets by \citet{savarese2019recurrent}, alternating templates allow odd and even residual steps to specialize in complementary feature transformations rather than forcing a single repeated mapping. While we intentionally restricted all continuation arms to a fixed one-epoch budget to isolate rapid structural recoverability from a warm checkpoint, testing whether a longer 5--10 epoch continuation schedule—or full 88-epoch training from scratch—closes the remaining $1.84$-point gap completely is a natural direction for future work. Finally, extending Stage~6 with a fifth recurrent pass ($[B_1, A, B, A, B]$, $+25\%$ stage compute and $+1,536$ BatchNorm parameters) yields $70.830\%$ vs.\ $70.916\%$ for the 4-block $[B_1, A, B, A]$ design (paired difference $-0.086$ points, 95\% CI $[-0.320, +0.148]$), indicating that four blocks already saturate Stage~6 representational depth at $224\times 224$ resolution.

\section{Scope, Limitations, and Conclusion}
\label{sec:conclusion}

\paragraph{Methodological Scope.}
Our B0/B1/B2 scale baselines ($72.86\%/74.91\%/75.95\%$) use an 88-epoch ReLU6/SGD recipe rather than the 350--400-epoch Swish/RMSProp protocol of \citet{tan2019efficientnet}, and differ in batch size across scales; they serve as controlled checkpoint anchors rather than a cross-scale law. Separate one-epoch CPU architecture pilots (including a differentiable TensorFlow reference formulation of DCNv4~[\citetalias{xiong2024dcnv4}] at $28.71\%$ FP32 and $28.69\%$ under selective LiteRT INT8/FP32 PTQ) verified operator portability and export pipelines rather than converged accuracy. Both the matched 88-epoch Sparse Attention pair ($+0.954$ pp) and the one-epoch GPU shared-repeat continuations ($70.92\%$) are controlled single-seed studies; in particular, the shared-repeat experiments evaluate rapid warm-start recovery over a fixed one-epoch budget rather than multi-epoch (5--10 epoch) fine-tuning or full from-scratch convergence.

\paragraph{Conclusion.}
Structural efficiency in compact vision backbones depends on respecting the functional hierarchy of convolutional stages. Heavy cross-scale feature pyramids (BiFPN) add overhead without beating a clean MBConv trunk in classification, whereas lightweight zero-gated self-attention adapters at $P_3/P_4/P_5$ naturally concentrate their activity in deep semantic stages ($66.3\%$ RMS at $P_5$) and deliver a verified $+0.954$-point matched accuracy gain ($73.88\%$ vs.\ $72.92\%$; $74.02\%$ standalone). Simultaneously, in wide deep stages (Stage~6), alternating two shared block templates ($[B_1, A, B, A]$) with pass-specific BatchNorm statistics cuts $10.9\%$ of total model parameters while recovering to $70.92\%$ in a single continuation epoch—providing a practical blueprint for compact, parameter-efficient hybrid vision architectures.

\makeatletter
\if@anonymous
\paragraph{Code and Weights Availability.}
Trained checkpoints, reference model-construction modules, and SHA-256 verification manifests will be released publicly upon publication.
\else
\paragraph{Code and Weights Availability.}
Trained checkpoints, reference model-construction modules, and SHA-256 verification manifests for this study (arXiv:2611.23446) are publicly available at \url{https://huggingface.co/meatloafer/efficientnet-b0-stage-attention} (companion sub-4-bit quantization study: arXiv:2611.23445, \url{https://huggingface.co/meatloafer/efficientnet-b0-sub4bit-mixed-precision}).
\fi
\makeatother

\bibliographystyle{plainnat}
\bibliography{references}

\end{document}